\documentclass[11pt,a4paper]{article}

\usepackage[T1]{fontenc}
\usepackage[utf8]{inputenc}
\usepackage[margin=1in]{geometry}
\usepackage{graphicx}
\usepackage{amsmath}
\usepackage{natbib}
\usepackage{microtype}
\usepackage{subcaption}
\usepackage[onehalfspacing]{setspace}
\usepackage{textcomp}
\usepackage{tipa}
\usepackage{xcolor}
\usepackage[hidelinks]{hyperref}

\makeatletter

\newcounter{panel}[figure]
\renewcommand{\thepanel}{\alph{panel}} 
\newcommand{\paneltag}[1]{%
  \refstepcounter{panel}%
  \rlap{\hspace*{-0.4em}\raisebox{0.3em}{\textbf{(\thepanel)}}}%
  \label{#1}%
}

\let\subref\ref
\makeatother

\begin{document}
\title{Encoding Tactile Stimuli for Braille Recognition with Organoids}
\author{%
Tianyi Liu\textsuperscript{1}, Hemma Philamore\textsuperscript{1}, and Benjamin Ward-Cherrier\textsuperscript{1,*}\\[0.5em]
\small \textsuperscript{1}School of Engineering Mathematics and Technology, University of Bristol, Bristol, United Kingdom\\
\small Corresponding author: Benjamin Ward-Cherrier (\texttt{b.ward-cherrier@bristol.ac.uk})
}
\date{}
\maketitle

\begin{abstract}
This study proposes a transferable encoding strategy that maps tactile sensor data to electrical stimulation patterns, enabling neural organoids to perform an open-loop artificial tactile Braille classification task. Human forebrain organoids cultured on a low-density microelectrode array (MEA) are systematically stimulated to characterize the relationship between electrical stimulation parameters (number of pulse, phase amplitude, phase duration, and trigger delay) and organoid responses, measured as spike activity and spatial displacement of the center of activity. Implemented on event-based tactile inputs recorded from the Evetac sensor, our system achieved an average Braille letter classification accuracy of 61\% with a single organoid, which increased to 83\% when responses from a three-organoid ensemble were combined. Additionally, the multi-organoid configuration demonstrated enhanced robustness against various types of artificially introduced noise. This research demonstrates the potential of organoids as low-power, adaptive bio-hybrid computational elements and provides a foundational encoding framework for future scalable bio-hybrid computing architectures.
\end{abstract}

\noindent\textbf{Keywords:} neural organoids, organoid intelligence, bio-hybrid computing, multiparameter encoding, open-loop tactile classification, braille classification, low-density MEA
\vspace{1em}

\section{Introduction}
Silicon-based neuromorphic processors  achieve notable energy savings when compared to traditional Von Neumann computing systems. Systems such as IBM's TrueNorth~\citep{akopyan2015truenorth} or Intel's Loihi~\citep{davies2018loihi}  operate in the milliwatt range by fusing memory and computation and firing only when events occur. However, when engineers scale these chips toward cortex-level complexity, projected power climbs into the kilowatt-megawatt bracket, whereas the human brain performs comparable computation on roughly 20 watts~\citep{mehonic2022brain}.

Biocomputing has been proposed as one route to address energy limitations in silicon-based systems by using brain organoids grown on multielectrode arrays.
These living networks have 3-D architecture and genuine synapses that reorganize and learn while consuming orders of magnitude less energy~\citep{perera2025wet}. Such capabilities emerge only when three layers operate as one stack: wetware (the organoid), hardware (microfluidic culture, MEAs, optical, and bio-CMOS sensors that keep the tissue alive and observable) and software (millisecond closed-loop algorithms that read, stimulate, and adapt the network). When these layers are co-designed, organoids shift from spontaneous activity to purposeful information processing~\citep{kagan2025harnessing}.

While some research primarily focuses on exploring internal neural dynamics and structure-function relationships within organoid networks, such as drug-induced synaptic modulation~\citep{shin20213d}, critical-state sensory conditioning~\citep{habibollahi2023critical}, and spontaneous functional network formation~\citep{sharf2022functional}, a growing body of work treats neural organoids as computing elements within artificial intelligence frameworks, a field termed Organoid Intelligence (OI)~\citep{smirnova2023organoid}.
OI leverages the intrinsic plasticity and extremely low power consumption of brain organoids, offering significant potential for creating energy-efficient computing platforms, particularly in robotics and autonomous systems.

Previous studies have demonstrated computational capabilities in a range of biological neural systems, including ex vivo circuits, cultured neuronal networks, and neural organoids.
Direction-selective ganglion-cell ensembles not only classified the direction of a moving bright bar but also exploited stimulus-dependent noise correlations to sharpen that estimate~\citep{franke2016structures}. 
Cortical organoids have acted as “living reservoir” computers that steadily improve at decoding Japanese vowels and chaotic time-series after characterization of their nonlinear dynamics~\citep{cai2023brain}.
In the DishBrain Pong paradigm, closed-loop interaction enabled the organoid to steadily improve its performance in the virtual game, indicating that the network had learned the task objective.
Cultured neurons shaped into a two-input/one-output pathway on a micro-electrode chip behave as a living logic gate, reliably delivering NAND or OR outputs and demonstrating that basic digital operations can run on biological tissue~\citep{kuchler2025engineered}. 
George et al. achieved BNN-ANN closed-loop control by bidirectionally linking a neonatal rat spinal CPG slice with an FPGA-based spiking ANN, restoring rhythmic bursts lost in isolation~\citep{george2020plasticity}.
Ades et al. built a closed-loop biohybrid  system in which mouse cortical neurons cultured on a multielectrode array drive a Shadow robotic hand with BioTac sensors~\citep{ades2024biohybrid}.
These findings substantiate organoids as feasible substrates for bio-hybrid intelligence, capable of learning, classification, and control in real-world robotic scenarios.

Fully realizing this potential in robotic applications requires effective communication among artificial sensory devices, organoid networks, and external processing-and-actuation systems.
This communication critically relies on suitable encoding and decoding methods. Encoding refers to transforming external sensory data or control signals into stimulation patterns that neural organoids can interpret, typically through electrical, optical, or chemical stimuli. Decoding involves translating the organoid responses into signals that external computational systems can process and interpret. Developing an efficient and transferable encoding-decoding framework is thus a prerequisite for enabling reliable interactions between robotic sensory systems and neural organoid computing platforms.
Despite the variation in organoid architectures and computational roles, multielectrode arrays (MEAs) have become the principal interface for stimulating and recording neural activity. High density MEAs have been successfully employed for closed-loop learning in planar cultures~\citep{kagan2022vitro} and reservoir computing in organoids~\citep{cai2023brain}, however, the cost and availability of such hardware remain limiting~\citep{xu2025multi}. Lower density layouts therefore continue to see extensive use, such as an implementation that positions each human forebrain spheroid above eight stainless-steel electrodes in an air-liquid-interface array and delivers remote recordings through the Neuroplatform~\citep{jordan2024open}. Using this same setup, we provide an open-loop platform for evaluating tactile-to-organoid encoding and decoding on a low-density MEA.

Prior studies in OI and biohybrid platforms explore strategies that encode stimuli and decode neural responses across varied tasks and organoid preparations.
Encoding implementations vary with task demands. Franke et al. employ natural stimuli, sweeping a bright bar across an ex-vivo retina allowing each ganglion cell to convert motion into its innate spike-count tuning, but limiting direct control over stimulus features~\citep{franke2016structures}. 
Rate-place coding across eight electrodes conveyed ball position to DishBrain cultures, enabling intuitive spatial representations suitable for closed-loop motor learning~\citep{kagan2022vitro}. 
Brainoware experiments use flexible spatiotemporal encodings tailored specifically to the input type: acoustic features are mapped to electrode location and stimulation amplitude, while scalar time-series are represented as amplitude-scaled pulses delivered at fixed intervals, balancing complexity and adaptability for distinct temporal inputs~\citep{cai2023brain}. 
The Shadow-Hand system encodes fingertip forces into slowly- and rapidly-adapting spike trains, using timing-only differences to deliver rapid change cues and continuous pressure feedback with a lightweight real-time Izhikevich model~\citep{ades2024biohybrid}. 
Address-event representation (AER) streams on the GAIA array encode space in the electrode address and time in the timing of each event, thus enabling seamless integration with neuromorphic hardware~\citep{cartiglia20244096}. Binary values have been represented in hippocampal micro-networks by toggling pulse amplitude or rate on two input lines, but was limited in representing richer analog or multidimensional signals~\citep{kuchler2025engineered}.
Decoding usually relies on straightforward spike-based statistics, most often the per-electrode spike count accumulated within a brief post-stimulus window. Alternative metrics such as time-to-first-spike and mean inter-spike interval are also employed~\citep{kuchler2025engineered}. 
Across their diverse aims, these studies consistently rely on application-specific encoding strategies.

In this work, we propose an encoding strategy for event based tactile inputs that extracts region wise spatial, temporal, and intensity related descriptors and maps them into stimulation parameters compatible with low-density MEAs.
Using an event-driven tactile sensor to capture Braille stimuli, we map sensory data into multi-channel electrical stimulation patterns. 
We demonstrate the feasibility and effectiveness of our method through a tactile Braille letter-classification task. 
On the remotely accessible FinalSpark Neuroplatform~\citep{jordan2024open}, tactile information from a neuromorphic sensor is delivered to human cortical organoids cultured on eight-electrode MEAs, and the decoded organoid activity is subsequently classified with a support vector machine (SVM)~\citep{cortes1995support}.
Our experiments show that neural organoids can robustly classify tactile patterns, and that employing multiple organoids in parallel improves classification accuracy and robustness, highlighting the promise of scalable biohybrid computing architectures. 
Our main contributions in this work are:

\begin{itemize}
    \item We establish a quantitative stimulus-response profile for human forebrain organoids on an eight-electrode low-density MEA, relating stimulation parameters to spike magnitude, timing, and spatial distribution.
    \item We present a tactile-to-organoid encoding and decoding framework that converts event-based tactile data into MEA-compatible electrical stimulation patterns while preserving spatial, temporal, and intensity-related tactile cues.
    \item We demonstrate Braille character classification from organoid responses, reaching 83\% accuracy with a three-organoid ensemble and improving robustness to Gaussian, uniform, missing-data, and outlier perturbations compared with single-organoid decoding and direct decoding from the delivered stimulation pattern.
\end{itemize}

\section{Materials and Methods}
\label{sec:Materials_and_Methods}

\subsection{Organoid and Recording Platform}
\label{subsec:Organoid_and_Recording_Platform}

\begin{figure}[htbp]
    \centering 
    \includegraphics[width=0.9\textwidth]{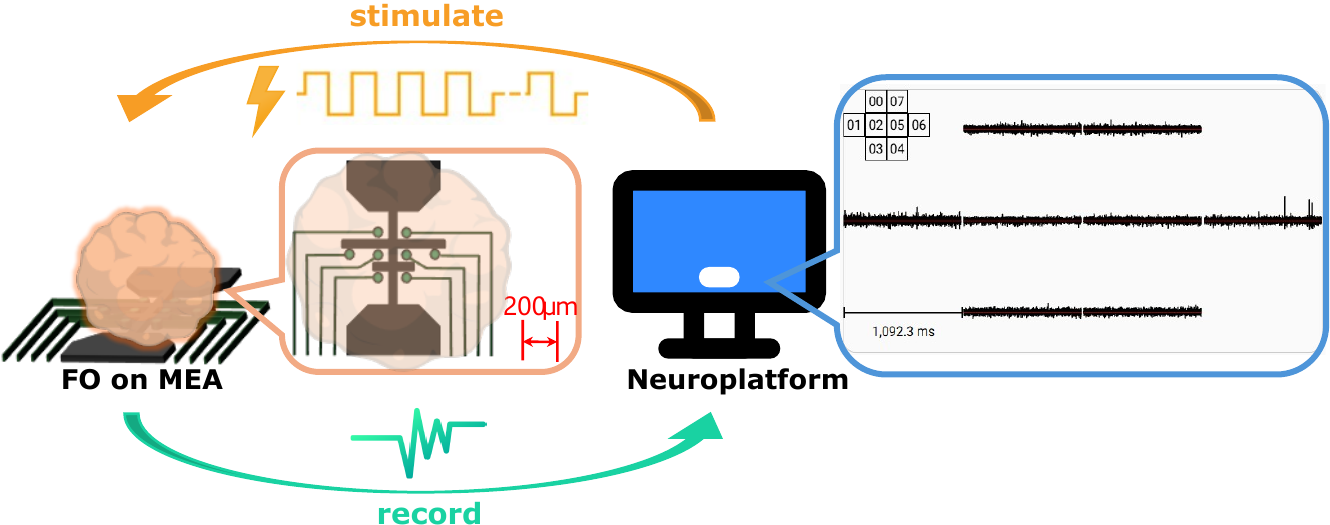} %
    \caption{Interaction between a forebrain organoid (FO) cultured on a multi-electrode array (MEA) and the Neuroplatform. 
    The 8 recording electrodes are arranged with 200 $\mu$m centre-to-centre spacing, providing a distributed interface across the organoid.
    Neural activity is recorded from the FO and electrical stimulation is delivered via a remotely accessible software interface.}
    \label{fig:finalspark} 
\end{figure}

All our experiments were conducted using FinalSpark's  NeuroPlatform~\citep{jordan2024open}, providing access to organoids over a cloud-based system, as shown in Figure~\ref{fig:finalspark}.
In this setup, forebrain organoids are cultured to approximately 500 $\mu$m in diameter~\citep{govindan2021mass} and contain mature neurons, astrocytes, and oligodendrocytes displaying robust network activity. 
Maintained under continuous perfusion, they can remain electrophysiologically active for up to approximately 100 days under favourable conditions~\citep{jordan2024open}.

The platform  consists of a remotely accessible electrophysiology system integrating four custom air-liquid interface MEAs, each accommodating four organoids with eight electrodes per organoid for a total of 32 channels per MEA. 
As illustrated in Figure~\ref{fig:finalspark} and Figure~\ref{fig:real_setup}a, each forebrain organoid is positioned on an eight-electrode interface with 200~$\mu$m centre-to-centre electrode spacing, and four organoids are accommodated on each MEA, giving 32 channels in total \citep{jordan2024open}. 
Voltage signals are acquired via Intan RHS headstages sampling at 30 kHz with 16-bit resolution. 
All hardware functions are accessible through a Python API and Jupyter notebook interface, with data continuously archived in an InfluxDB time-series database which can be used for downstream analysis.
In this study, the Neuroplatform was accessed remotely for stimulation and recording, whereas tactile sensing and data preprocessing were performed locally before stimulation was executed.

\subsection{The Impact of Electrode Stimulation Parameters on Organoid Responses}
\label{subsec:Impact_of_Electrode_Stimulation_Parameters}

Across the three experiments, we tested organoid responses to stimulation parameters representing the stimulus’s temporal, spatial, and intensity dimensions, laying the groundwork for an encoding strategy that converts tactile-sensor outputs into electrical patterns.

\begin{figure}[htbp]
    \centering 
    \includegraphics[width=0.8\textwidth]{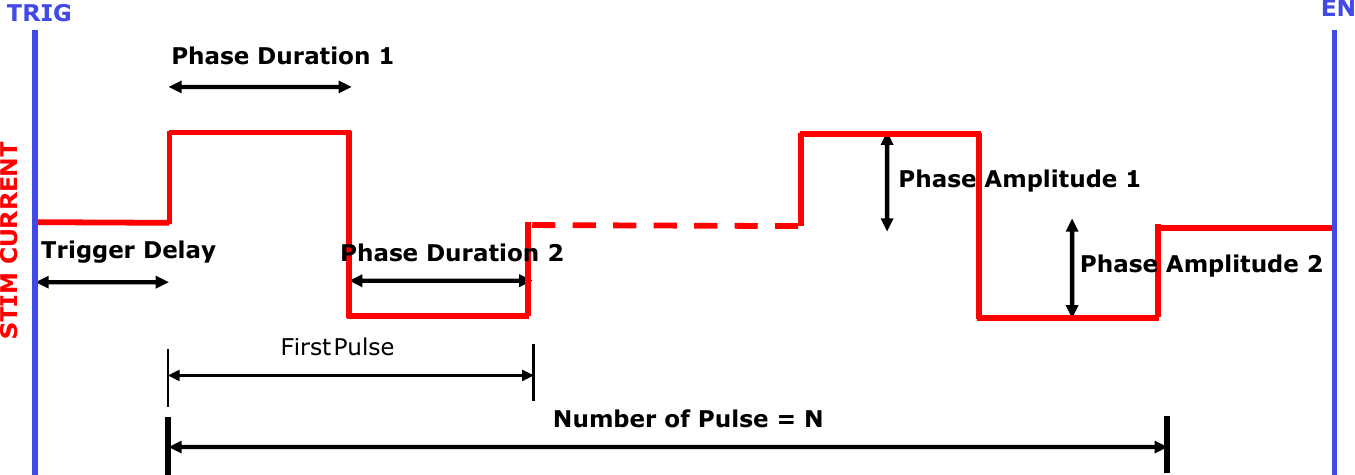} %
    \caption{Illustration of the charge balanced biphasic current waveform used in this study and its adjustable parameters on the Neuroplatform.}
    \label{fig:param} 
\end{figure}

As shown in Figure~\ref{fig:param}, electrical stimulation on the Neuroplatform is parameterized through several key variables~\citep{jordan2024open}. The “shape” of the current waveform may be biphasic, biphasic with an interphase delay, or triphasic. 
A biphasic waveform consists of a positive phase followed by a negative phase, each specified by an amplitude and a duration. A triphasic waveform consists of three phases with alternating polarity.
Each phase is specified by an amplitude and a duration: Amplitude 1 and Duration 1 for the positive phase, Amplitude 2 and Duration 2 for the negative phase. 
The polarity of the waveform can be inverted to begin with either a negative or a positive phase. Stimulation can be triggered as a single pulse, as a sequence defined in a trigger table, and can be directed to any subset of the 32 electrodes. The trigger delay parameter controls the onset timing of the first pulse, while the number of pulses parameter specifies how many pulses are delivered in each sequence.
Charge balanced biphasic stimulation is widely used because its primary objective is to keep the electrode potential within a safe range, thereby reducing irreversible electrochemical reactions that can degrade the electrode or damage tissue~\citep{cogan2008neural}. Biphasic current stimulation has previously been shown to elicit spikes and modulate network activity in forebrain organoids on the FinalSpark Neuroplatform~\citep{jordan2024open}.
In all experiments, we chose a positive first and negative second biphasic current waveform because it statistically produces the lowest neural activation thresholds while safeguarding both the electrode and surrounding cells over the long term~\citep{wagenaar2004effective}. 
To improve the lifetime of both the organoid and the electrodes, we set Phase Amplitude 1 = Phase Amplitude 2 and Phase Duration 1 = Phase Duration 2.
We used the Neuroplatform default interphase delay of 0 $\mu$s between the two phases, and default pulse train period of 10000 $\mu$s for pulse trains.

\subsubsection{Organoid Responses Under Different Stimulation Parameters}

In this experiment, we map the effective stimulation range by testing whether electrical input produces significant departures from spontaneous activity and whether different parameter values elicit statistically separable response profiles. Values that meet both criteria become candidate parameters for the encoding scheme. Finally, we assess whether parameter–response relationships are consistent across electrodes to determine if encoding must be tailored for each electrode.

The experiment involves stimulating each electrode within the same organoid respectively, recording data from all electrodes and averaging across them to observe the overall effect of stimulation parameters on the organoid output. Stimuli are delivered with a 2 s inter‐stimulus interval to allow sufficient recovery and avoid temporal summation. 
Here, one stimulus refers to a single trial consisting of a pulse train with 1 to 10 pulses, the interpulse timing within the train is set by the platform default 10000 $\mu$s pulse train period.
We record the number of threshold-based events generated across all channels from 10 to 210 ms relative to trigger onset, excluding the first 10 ms to suppress immediate stimulation artifacts~\citep{finalspark_closedloop_doc}, and compute the mean over 10 trials.
We test various numbers of pulses (0 to 10), phase amplitudes (0 to 20 $\mu$A), phase durations (0 to 200 $\mu$s), and trigger delays (0 to 4000 $\mu$s). A parameter value of 0 indicates spontaneous activity recorded under the unstimulated (spontaneous) condition. During each individual sweep, the remaining parameters are kept at their default values: a single pulse, 4 $\mu$A phase amplitude, and a 100 $\mu$s phase duration.
These base parameters were selected to provide a simple and interpretable reference condition for one factor sweeps. We used a single pulse as the minimal stimulation form, minimizing temporal summation effects and prolonged network activation. We set the phase amplitude to 4 $\mu$A because lower amplitudes were typically subthreshold in our recordings, while higher values risked driving near-saturated responses, and chose a 100 $\mu$s phase duration as a mid-range value within the sweep to avoid boundary effects.

\subsubsection{Temporal Dynamics of Organoid Responses}

We then investigate how long the effect of a single stimulation persists, to determine the appropriate time window for each sample in the Braille experiment. All channels of the same organoid are simultaneously stimulated with identical parameters; the average spike count across all channels is recorded over a 600 ms window around the stimulus. 

Stimulation parameters explored in this experiment are as follows: numbers of pulses (1 - 10 pulses), amplitude (2 - 20 $\mu$A), and phase duration (25 - 200 $\mu$s).
We sweep one parameter at a time through its range while keeping the other two fixed at their default values of a single pulse, 4 $\mu$A, and 100 $\mu$s. Each parameter condition is repeated 10 times with a 2~s inter-stimulus interval to allow recovery. We record spike counts from 100 ms before stimulation to 500 ms after stimulation across all channels and average over 10 trials.

\subsubsection{Spatial Dynamics of Organoid Responses}

We adopted the method of Jordan et al.~\citep{jordan2024open} to estimate the spatial center of activity (CA) after each stimulus, thereby quantifying shifts in overall network firing. The CA is the weighted centroid of spikes recorded on the organoid’s eight electrodes.
For a 500 ms analysis window, with spike count $F_k$ and normalized coordinates $(X_k,Y_k)$ at electrode $k$, the CA vector $\mathbf{C}$ is:

\begin{equation}
\mathbf{C} = \frac{\sum_{k=1}^{8} F_k \cdot (X_k, Y_k)}{\sum_{k=1}^{8} F_k}
\end{equation}

We first recorded the CA during spontaneous activity to establish a baseline.  
For each stimulation condition, we recorded spike data from all electrodes within a 500~ms post-stimulation window. We systematically tested and recorded the organoid responses by stimulating each electrode with varying numbers of pulses (2, 4, 6, 8, and 10), phase amplitudes (4, 8, 12, 16, and 20 $\mu$A), and phase durations (50, 100, 150, 200, and 250 $\mu$s), while the other electrodes were not stimulated. Spike counts were still recorded from all eight electrodes and used to compute the CA for each trial.
The default stimulation parameters are set to a pulse count of 5, a phase amplitude of 10 $\mu$A, and a phase duration of 150 $\mu$s, corresponding to the midpoints of their respective parameter ranges. 
This keeps the reference condition away from both near-threshold and near-saturation regimes, so that changes induced by the swept parameter are more readily observable and less affected by boundary effects.
Each condition is repeated 100 times with a 500~ms inter-stimulus interval.

The CA displacement relative to baseline shows how population firing migrates within the organoid. By analysing these displacements we test whether assigning different regions of the tactile sensor to distinct electrodes preserves spatial information and whether varying stimulus magnitudes elicits distinguishable network responses.

\subsection{Applying Organoid Intelligence to Braille Classification}

Earlier stimulation tests showed that variations in pulse amplitude, width and count elicit distinct firing patterns in the organoid (see Section~\ref{subsec:Effects_of_Stimulation_Parameters}). Building on these results, we design an encoding pipeline that converts event based tactile signals into electrical stimulation patterns and applies straightforward machine learning classifiers to the resulting organoid activity. 

\begin{figure}[htbp]
    \centering 
    \includegraphics[width=1\textwidth]{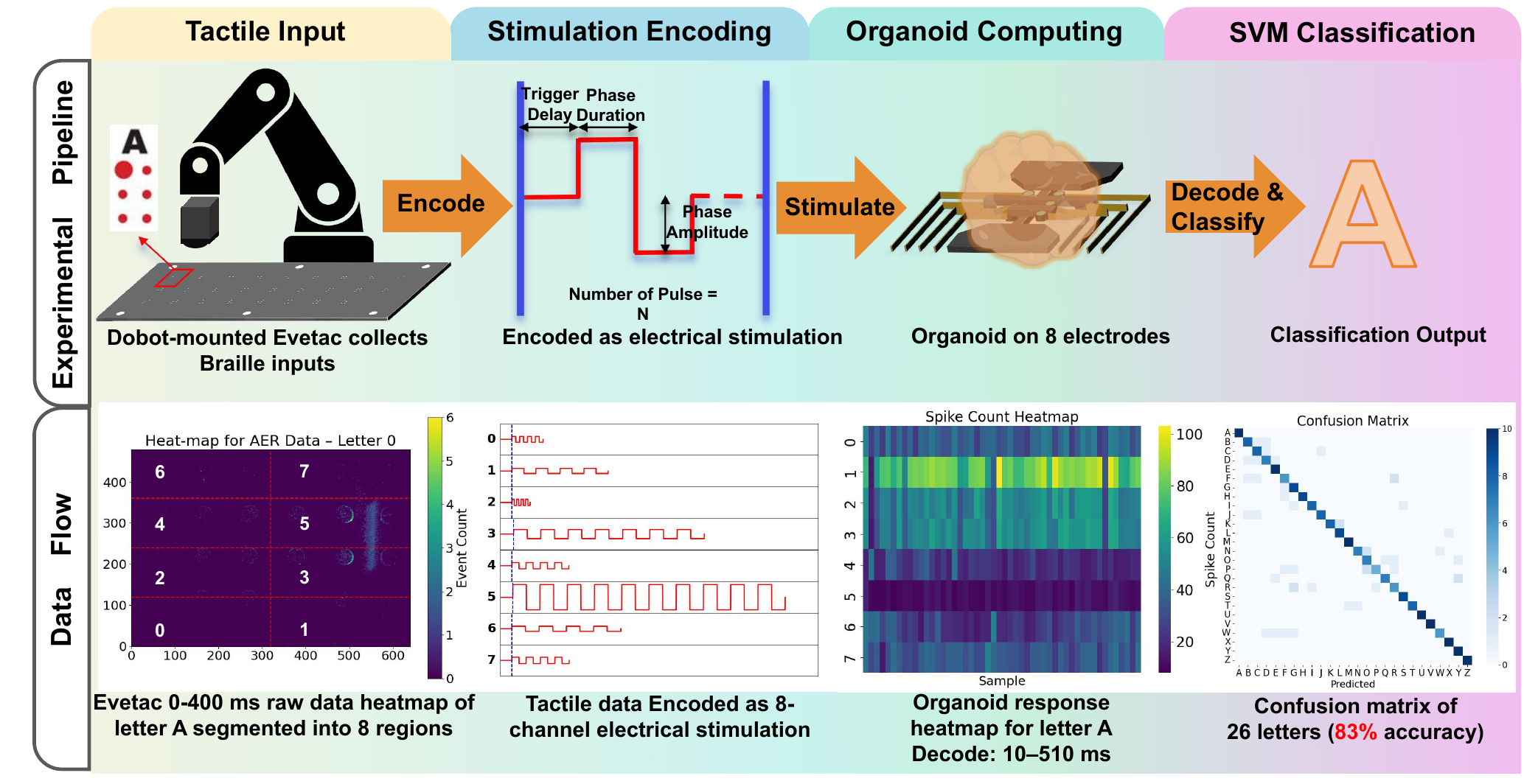} %
    \caption{Experimental procedure for collecting Braille using a tactile sensor, encoding it into electrical stimulation, delivering it to organoids, and classifying 26 Braille letters based on organoid outputs. The first row outlines the experimental steps, and the second row shows the type of data associated with each step. The tactile input is recorded over a 400 ms window. For organoid decoding, neural activity is recorded from 0 to 510 ms relative to trigger onset. The first 10 ms are excluded to suppress immediate stimulation artifacts. Threshold-based event counts from 10 to 510 ms are used for classification.}
    \label{fig:exp_workflow} 
\end{figure}

\subsubsection{Experimental Setup and Data Collection}

\begin{figure}[htbp]
    \centering 
    \includegraphics[width=1\textwidth]{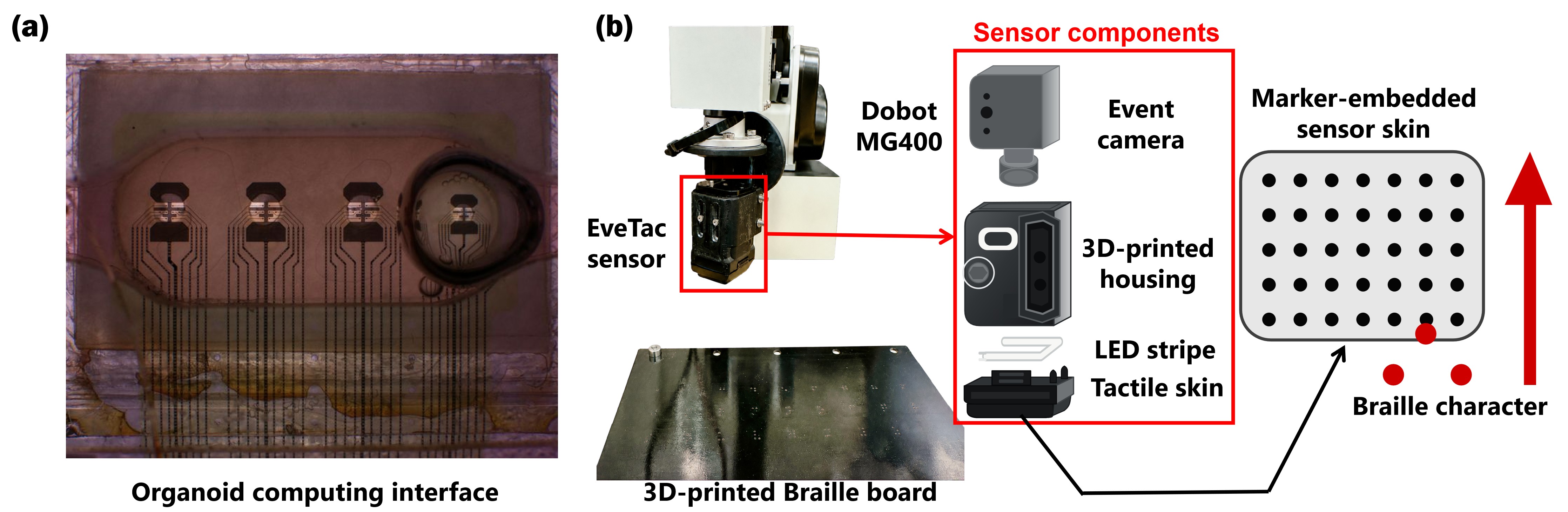} %
    \caption{Tactile sensing and organoid computing interfaces. (a) FinalSpark Neuroplatform MEA used in this study. The 32 electrodes are arranged in four groups of eight, with one forebrain organoid positioned over each group~\citep{jordan2024open}. (b) Evetac acquisition setup and sensing principle. From left to right: the Evetac sensor mounted on the Dobot MG400 robotic arm above the 3D-printed Braille board; the sensor components, including an event camera, 3D-printed housing, LED stripe, and marker-embedded tactile skin~\citep{funk2024evetac}; and the Braille character sliding across the marker-embedded skin in the direction indicated by the red arrow.}
    \label{fig:real_setup} 
\end{figure}

An Evetac optical tactile sensor~\citep{funk2024evetac} was mounted on the end effector of a four-axis Dobot MG400 robotic arm~\citep{dobot}, as shown in Figure~\ref{fig:real_setup}b. The sensor contains a soft silicone gel embossed with a regular grid of black markers, an internal LED ring for uniform illumination and a DVXplorer Mini event camera housed in a 3d-printed enclosure. Gel indentation displaces the markers, the accompanying motion and shadow pattern modulate local brightness, generating asynchronous pixel events that the camera exports every millisecond. Each event carries its pixel address, polarity and a microsecond time stamp and is streamed in Address-Event Representation (AER) format~\citep{boahen2002point}. 
Evetac resolves vibrations up to 498 Hz and captures shear forces and incipient slip~\citep{funk2024evetac}, making it suitable for fast Braille exploration.
Tactile stimuli are delivered with a 3D-printed Braille board that held the 26 English letters arranged in three rows of nine, nine and eight characters. The plate measures 23×14~cm, and adjacent letters are separated by 2.2~cm.

The overall pipeline is as displayed in Figure~\ref{fig:exp_workflow}. First, Braille characters are sampled using the Evetac sensor~\citep{funk2024evetac}: for each of the 26 Braille letters, the Evetac is lowered to a given depth (0, 0.1, 0.2, 0.3, 0.4 mm) and slid horizontally across the  Braille character for 10 trials, yielding 1300 recordings in total. 
Depth 0 denotes the initial shallow indentation upon contact.
Next, we encode this AER data into stimulation patterns for the organoid's 8 electrodes (as detailed in Section~\ref{sec:methodology_encoding}).  
Then stimulation patterns for the different Braille characters are delivered to the organoid and the organoid's response is read out as an 8‐dimensional response vector of spike counts across the organoid’s eight channels.
For each sample, neural activity is recorded from 0 to 510 ms relative to trigger onset, the first 10 ms are excluded to suppress immediate stimulation artifacts~\citep{finalspark_closedloop_doc}, and counts from 10 to 510 ms are used for classification. 
Finally, appropriate classifiers are applied to distinguish among the 26 Braille letters based on organoid responses. We test KNN~\citep{cover1967nearest}, SVM~\citep{cortes1995support}, and random forest~\citep{breiman2001random} classifiers, using the same feature representation and cross-validation protocol. SVM was selected for the main analyses because it achieved the highest mean cross-validated classification accuracy among the tested classifiers. This choice is also suitable for the relatively low-dimensional spike-count features used here, where margin-based classifiers can perform well without requiring large training sets.

\subsubsection{Electrical Encoding of Tactile Event Features}
\label{sec:methodology_encoding}

\begin{figure}[htbp]
    \centering 
    \includegraphics[width=1\textwidth]{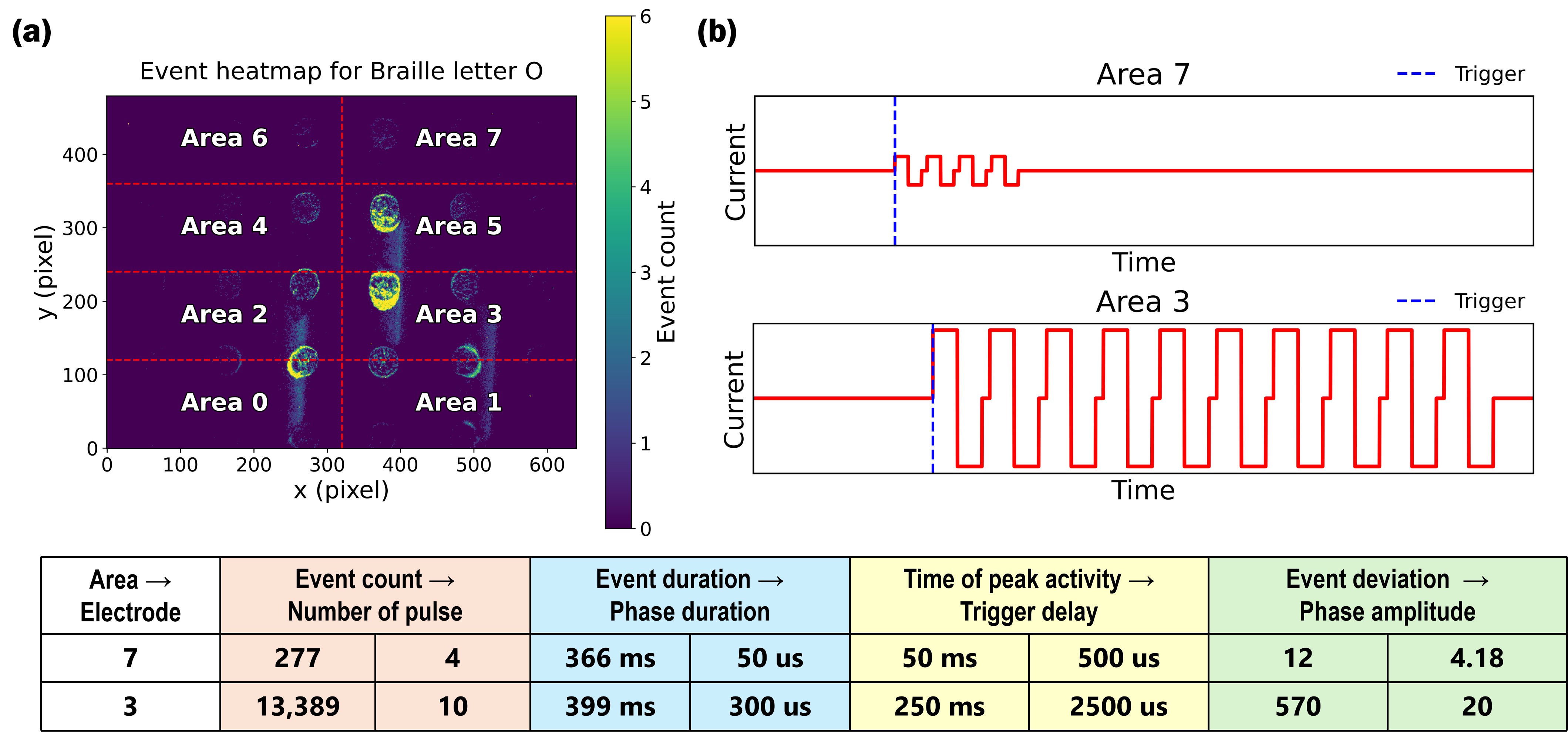} %
    \caption{Illustration of the encoding strategy. (a) The output of the neuromorphic tactile sensor for Braille letter O in the form of an event heatmap, where the colorbar indicates the event count and brighter colors represent higher event counts. (b) The corresponding electrical stimulation waveforms for Areas 7 and 3, shown schematically. The table below presents an example of mapping sensor outputs in Areas 7 and 3 to electrical stimulation parameters.}
    \label{fig:encoding_eg} 
\end{figure}

We partition the Evetac output into eight spatial regions that correspond to the organoid’s eight channels and introduce an encoding that embeds spatiotemporal intensity information to convert tactile events into electrical stimulation.
From the AER data~\citep{chan2007aer}, four key features are extracted for each region:

\begin{itemize}
\item \textbf{Event count}: the total number of events in the region, representing overall tactile intensity.
Here, an “event” refers to a single AER pixel event with an address, polarity, and timestamp. The event count for a region is computed as the total number of raw AER events whose pixel addresses fall within that region, summed over the full 400 ms tactile recording.
\item \textbf{Event duration}: the time difference between the first and last events, representing the duration of tactile contact.
\item \textbf{Time of peak activity}: the midpoint time of the 100 ms window with the highest event count, indicating when the Braille character reaches that region (original data length was 400 ms).
\item \textbf{Event deviation}: the standard deviation of event counts across 100 ms windows, representing temporal variation in tactile intensity.
\end{itemize}

We select four key stimulation parameters for the organoids: number of pulses, phase duration, trigger delay, and phase amplitude, and all electrodes use the same encoding range. 
These parameters control the timing and strength of the stimulation delivered to the organoid.
For each AER recording, each extracted feature is linearly mapped to its corresponding stimulation parameter:

\begin{itemize}
\item \textbf{Number of pulses}: controls the number of pulses in each spike train to convey tactile strength; mapped to 4--10.
\item \textbf{Phase duration}: controls the duration of each pulse to convey tactile duration; mapped to 50--300 $\mu$s.
\item \textbf{Trigger delay}: controls the time to start the stimulation to convey when the tactile event begins; mapped to 0--4000 $\mu$s. 
\item \textbf{Phase amplitude}: controls pulse strength to convey temporal variation of tactile intensity; mapped to 4--20 $\mu$A.
\end{itemize}

We set a 1 s inter-sample interval on an onset-to-onset basis. For each sample, neural activity was recorded from 0 to 510 ms relative to trigger onset. The first 10 ms were excluded from analysis to suppress contamination from immediate stimulation artifacts~\citep{finalspark_closedloop_doc}, and threshold-based event counts from 10 to 510 ms were used as the 500 ms organoid output window, as shown in Figure~\ref{fig:encoding_eg} for a sample encoding of the letter~O.

\subsubsection{Organoid Decoding and Classification Under Noisy Conditions}
\label{subsubsec:Organoid_Decoding_and_Classification_Under_Noisy_Conditions}

We decode the spike patterns that emerge from the organoid’s own network dynamics after each stimulus. 
For each sample, the classifier input consisted of trigger-aligned, threshold-based event counts recorded separately on each electrode. Neural activity is recorded from 0 to 510 ms relative to trigger onset, the first 10 ms are excluded to suppress immediate stimulation artifacts~\citep{finalspark_closedloop_doc}, and counts from 10 to 510 ms are used as the 500 ms feature window. 
Since each organoid is connected to 8 electrodes, the resulting spike activity forms an 8-dimensional feature vector, which is then used by a support vector machine classifier to identify the Braille character. 
To study the effect of scale, we repeat the experiment with 3 organoids in parallel. The same 8-region encoding drives each organoid, we record all 8 channels from every organoid, and we concatenate them into a 24-dimensional feature vector before classification.

To assess robustness, we introduce artificial disturbances to the spike-count features while keeping the original tactile dataset unchanged. The tactile data are collected at 5 different depths, which already introduces inherent variability, the additional disturbances are applied on top of this. 
Specifically, we randomly select 4 of the 8 channel positions and modify the corresponding values for each organoid, resulting in 4 perturbed channels in the single-organoid condition and 12 perturbed channels in the three-organoid condition. 
We apply 4 disturbance models: additive Gaussian noise with a standard deviation equal to 40\% of the channel standard deviation, additive uniform noise within the same range, data loss where 40\% of the selected entries are set to zero, and outliers where 40\% of the selected entries are multiplied by 3. We then report support-vector machine accuracy under each model to quantify how single and triple organoid decoders tolerate channel interference. These models are not intended to cover every realistic sensor artifact, but to provide a basic stress test of whether the performance benefit of using multiple organoids remains under several common types of channel corruption.

\subsubsection{Direct Stimulation Baseline}

To provide a direct machine-learning baseline, we also applied the same SVM classifier directly to the delivered stimulus waveform, without using organoid responses. In this representation, each sample was fully specified by four encoded parameters on each of the eight stimulation channels, namely number of pulses, phase amplitude, phase duration, and trigger delay, resulting in a 32-dimensional stimulation vector. 
The same cross-validation protocol was used as in the organoid-based decoding experiments. For robustness analysis, we applied the same channel-corruption scheme as in Section~\ref{subsubsec:Organoid_Decoding_and_Classification_Under_Noisy_Conditions}: four of the eight channels were randomly selected, and the corresponding parameter blocks were modified using Gaussian noise, uniform noise, missing-data corruption, or outlier corruption.

\section{Results}

\subsection{The Impact of Electrode Stimulation Parameters on Organoid Responses}
\label{subsec:Effects_of_Stimulation_Parameters}

In this section, we examine how individual electrode stimulation parameters affect organoid responses. We quantify effects in three aspects: spike count, temporal dynamics, and spatial dynamics. Spike counts are analyzed through controlled parameter sweeps with trend visualizations to reveal overall patterns. Temporal dynamics are characterized from response profiles within a fixed readout window after stimulation. Spatial dynamics are assessed using a set of three metrics to evaluate displacement and the clustering and dispersion of organoid responses. The combined evidence guides the selection and mapping ranges of stimulation parameters for converting tactile inputs into electrical stimuli for the organoid system.

\subsubsection{Organoid Responses Under Different Stimulation Parameters}

\label{subsubsec:Spike_Under_Varying_Parameter}

\begin{figure}[htbp]
\centering

\begin{minipage}[t]{0.49\textwidth}
  \centering
  \paneltag{subfig:elec_param_NP}
  \includegraphics[width=\linewidth]{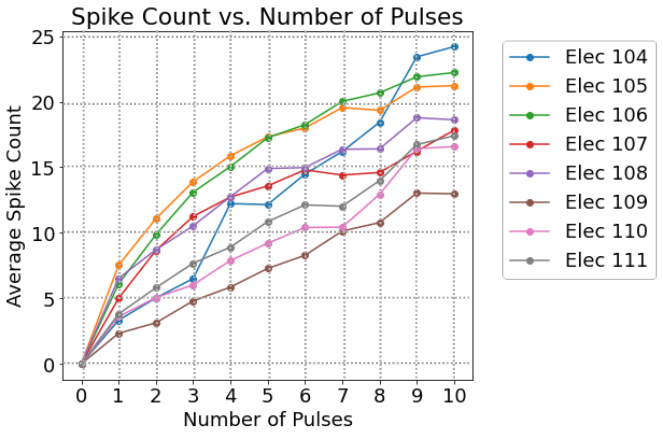}
\end{minipage}\hfill
\begin{minipage}[t]{0.49\textwidth}
  \centering
  \paneltag{subfig:elec_param_PA}
  \includegraphics[width=\linewidth]{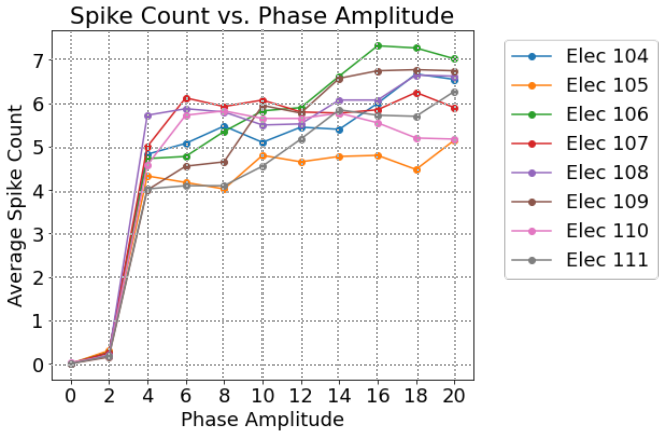}
\end{minipage}

\vspace{0.8\baselineskip}

\begin{minipage}[t]{0.49\textwidth}
  \centering
  \paneltag{subfig:elec_param_PD}
  \includegraphics[width=\linewidth]{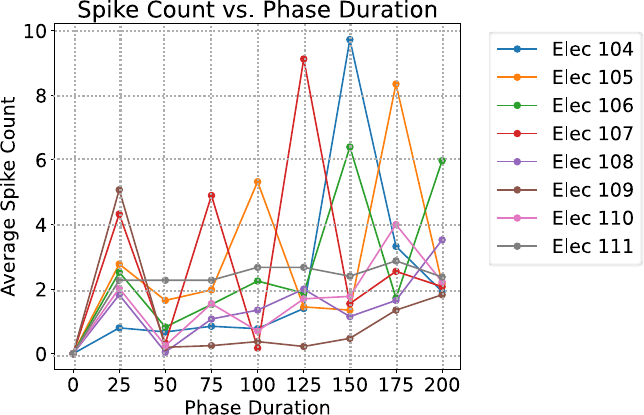}
\end{minipage}\hfill
\begin{minipage}[t]{0.49\textwidth}
  \centering
  \paneltag{subfig:elec_param_TD}
  \includegraphics[width=\linewidth]{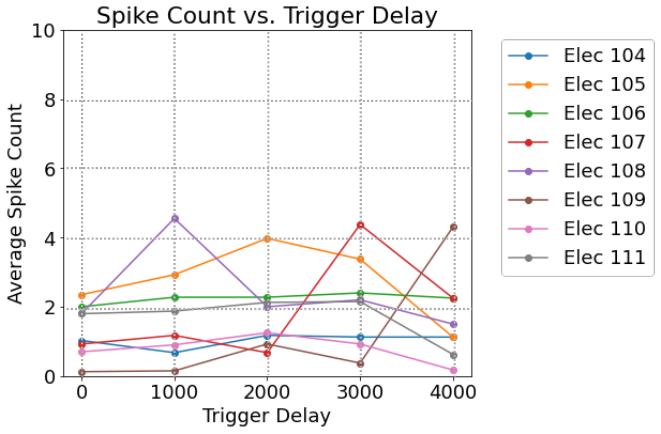}
\end{minipage}

\caption{Relationship between stimulation parameter values and the total spike count across all channels of the organoid.
(\ref{subfig:elec_param_NP}) Number of Pulses.
(\ref{subfig:elec_param_PA}) Phase Amplitude.
(\ref{subfig:elec_param_PD}) Phase Duration.
(\ref{subfig:elec_param_TD}) Trigger Delay.
The x-axis represents the parameter value, and the y-axis indicates the average spike count across all channels. Lines represent the index of the stimulated electrode.}
\label{fig:elec_param}
\end{figure}

This experiment tests how spike output varies when one stimulation parameter changes at a time, guiding the choice of safe and informative encoding ranges.
Stimulating each electrode in turn and recording from all eight channels reveals clear trends (Figure~\ref{fig:elec_param}). Spike count increases almost linearly as the number of pulses rises and shows a threshold for phase amplitude: values below 4 $\mu$A rarely evoke spikes, whereas higher amplitudes produce steadily larger responses. 
In this one-factor sweep, phase duration does not produce a consistent monotonic trend in spike count across the tested conditions, and trigger delay has no observable effect on spike count. 
Under the no stimulation condition, the spike count measured within the same 200 ms window is often close to zero. This reflects the combination of sparse spontaneous firing and the short counting window, rather than an absence of spontaneous activity.

Taken together, the observed response trends motivate the following encoding choices. The similarity across electrodes supports adopting a common encoding range for all channels. Phase amplitude is set above the activation threshold while remaining within safe limits. Pulse count is chosen from the region where spike output increases approximately linearly, enabling graded intensity. Although phase duration and trigger delay do not show a clear correlation with spike number, varying them produces distinct temporal patterns, so both are retained to encode timing information.

\subsubsection{Temporal Dynamics of Organoid Responses}

\begin{figure}[htbp]
    \centering

    \begin{minipage}[t]{0.34\textwidth}
        \centering
        \paneltag{subfig:param_time_NP}
        \includegraphics[width=\linewidth]{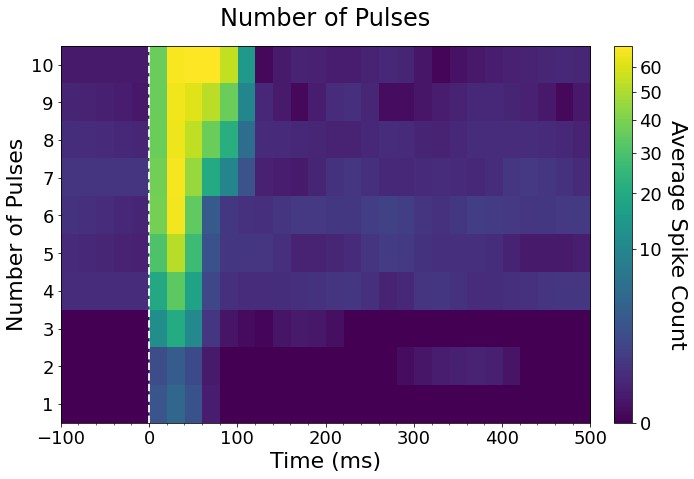}
    \end{minipage}\hfill
    \begin{minipage}[t]{0.33\textwidth}
        \centering
        \paneltag{subfig:param_time_PA}
        \includegraphics[width=\linewidth]{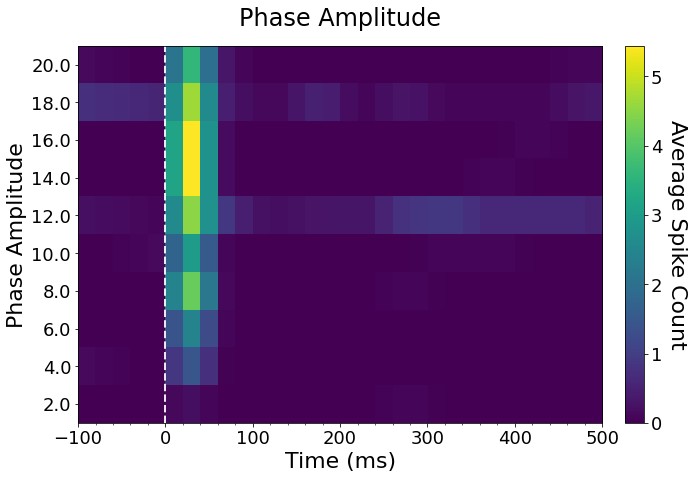}
    \end{minipage}\hfill
    \begin{minipage}[t]{0.33\textwidth}
        \centering
        \paneltag{subfig:param_time_PD}
        \includegraphics[width=\linewidth]{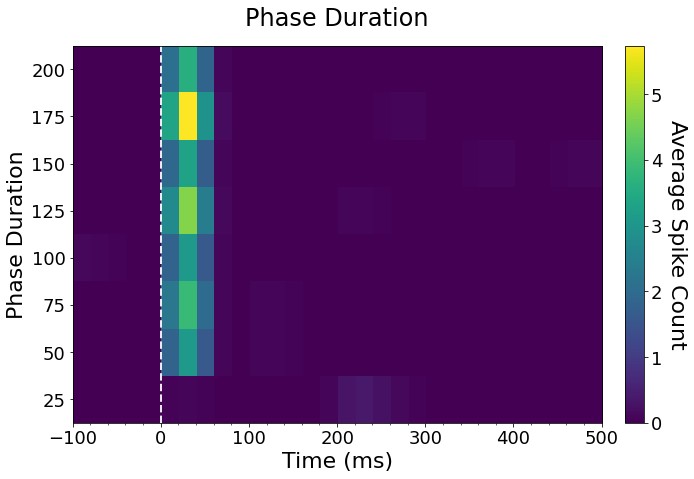}
    \end{minipage}
    \caption{Relationship between stimulation parameter value and duration of the stimulation effect. 
    (\subref{subfig:param_time_NP}) Number of Pulses.
    (\subref{subfig:param_time_PA}) Phase Amplitude.
    (\subref{subfig:param_time_PD}) Phase Duration.
    In each heatmap, the x-axis displays time from 100 ms before to 500 ms after stimulation, while the y-axis represents the varying parameter values. The color intensity indicates the average spike count of the organoid, with brighter colors representing higher activity levels. The vertical dashed line at t=0 indicates the stimulation onset.}
    \label{fig:param_time}
\end{figure}

This experiment measures how long each stimulus influences network firing.
In this experiment, we apply simultaneous stimulation across all electrodes with varying parameters and partition the recording into 20 ms windows, as shown in Figure~\ref{fig:param_time}. The dashed line marks stimulation onset.

When a single pulse is delivered, most responses return to the pre-stimulation activity level within 60 ms. 
Figure~7\ref{subfig:param_time_NP} shows that increasing the pulse count effectively prolongs the duration of the evoked activity, extending the response to approximately 100 ms at higher pulse counts.
Figure~7\ref{subfig:param_time_PA} again shows that phase amplitudes below 4 $\mu$A fail to evoke responses, whereas larger amplitudes do not uniformly increase spike count, they result in distinguishable intensities of activation within the post-stimulation window.
Even a 50 \textmu s pulse triggers a clear spike burst (Figure~7\ref{subfig:param_time_PD}), in line with earlier reports \citep{wagenaar2004effective}.
These findings motivate the use of an approximately 500 ms decoding duration in the Braille experiment. Figure~\ref{fig:param_time} shows the full trigger-aligned temporal profile, including the immediate post-trigger interval, to characterize the evolution of activity after stimulation. In the Braille classification pipeline, however, the first 10 ms after trigger onset are excluded to suppress immediate stimulation artifacts, and threshold-based event counts from 10 to 510 ms are used for decoding. 
The Neuroplatform’s adaptive threshold references each electrode’s baseline level, making sure a majority of spikes detected in this window arise from the stimulus.

\subsubsection{Spatial Dynamics of Organoid Responses}

\begin{figure}[htbp]
    \centering

    \begin{minipage}[t]{\textwidth}
        \centering
        \paneltag{subfig:CA_silhouette}
        \includegraphics[width=\linewidth]{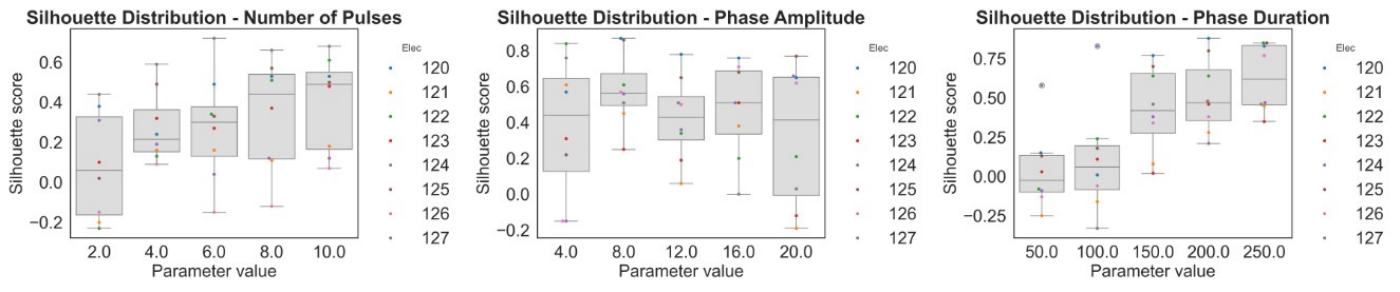}
    \end{minipage}

    \begin{minipage}[t]{\textwidth}
        \centering
        \paneltag{subfig:CA_dist}
        \includegraphics[width=\linewidth]{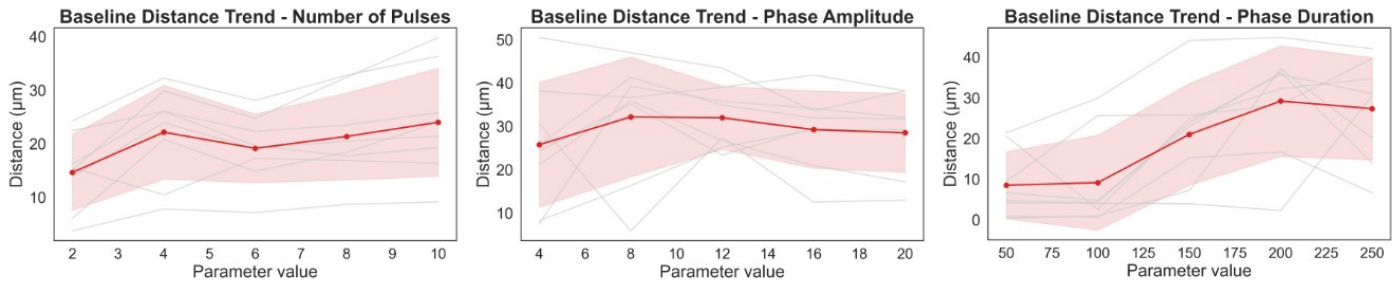}
    \end{minipage}

    \begin{minipage}[t]{\textwidth}
        \centering
        \paneltag{subfig:CA_spike}
        \includegraphics[width=\linewidth]{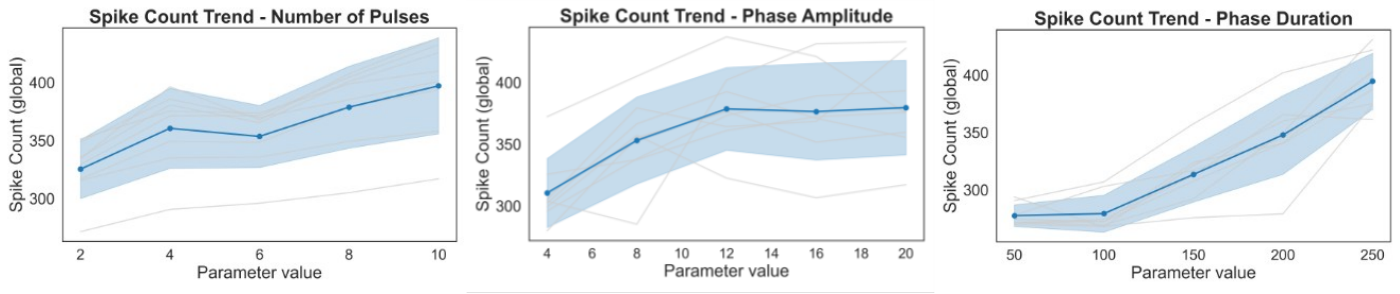}
    \end{minipage}

    \begin{minipage}[t]{\textwidth}
        \centering
        \paneltag{subfig:CA_radar}
        \includegraphics[width=\linewidth]{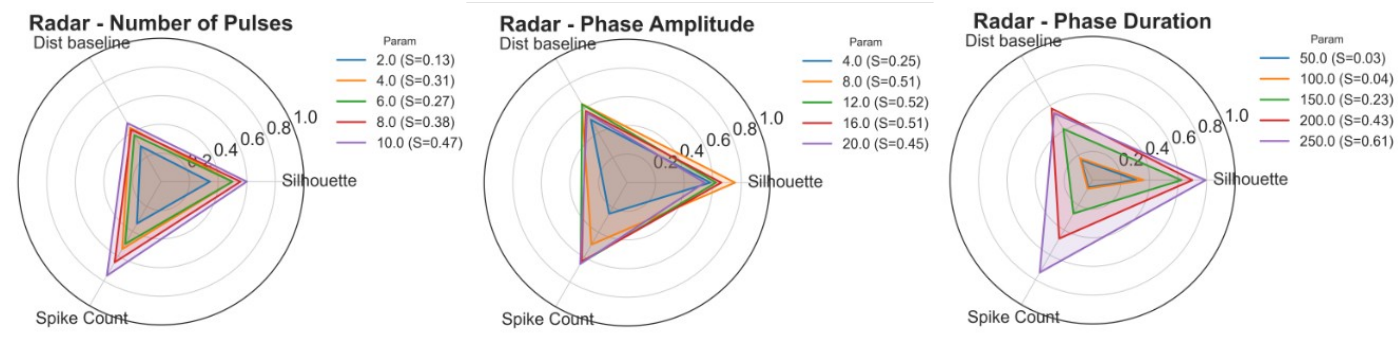}
    \end{minipage}

    \caption{From left to right the three columns correspond to Number of Pulses, Phase Amplitude, and Phase Duration.
    (\subref{subfig:CA_silhouette}) Silhouette Distributions. Box-and-swarm plots show cluster compactness/separation for each stimulating electrode.  
    (\subref{subfig:CA_dist}) Baseline-centroid Shift. Grey lines are per-electrode distances from the spontaneous centroid, the red line ± band is the array mean ± SD.  
    (\subref{subfig:CA_spike}) Global Spike Count. Analogous representation for total evoked spikes, with blue mean ± SD. 
    (\subref{subfig:CA_radar}) Radar Summaries. Three-axis polygons (Silhouette, baseline shift, spike count) condense the metrics for every parameter value, legend lists parameter and polygon area S.}
    \label{fig:CA_overall}
\end{figure}

This experiment assesses how stimulation parameters reshape the spatial pattern of activity across the organoid.
Each stimulus value was applied individually to every stimulating electrode; for each electrode-parameter pair we recorded 100 trials, yielding a 100-point activity-centre (CA) cluster. The 100 CA locations from one electrode at one setting form a cluster, points in the same cluster share the same electrode and parameter, whereas points in different clusters share the parameter but not the electrode. To characterise how a stimulus reshapes these clusters, we computed three quantitative metrics:
\begin{itemize}
    \item \textbf{Silhouette score}: Combines intra-cluster tightness with inter-cluster separation. A higher score indicates a CA cloud that is simultaneously compact and well isolated from those elicited by the other electrodes.
    \item \textbf{Centroid shift from baseline}: Euclidean distance between the stimulated cluster centroid and the spontaneous (no-stimulus) CA centroid. This captures the spatial displacement produced by the stimulus.
    \item \textbf{Global spike count}: Total spikes across the eight recording channels, averaged over the 100 repetitions. This reflects the overall population excitability evoked by the stimulus.
\end{itemize}

Figure~\ref{fig:CA_overall} congregates the three quantitative descriptors into aligned panels so that the influence of each stimulus family can be followed across electrodes and then integrated in Figure~8\ref{subfig:CA_radar}. In the radar plots each metric is first min-max normalised to 0-1 using the global extrema across all parameters and electrodes, then averaged over electrodes. Sharing the same (min, max) pair places every radius on a common percentile scale, producing unit-free polygons that can be compared directly between parameter values and across protocols.

With pulse count the upward trend is most evident between two and four pulses: the median Silhouette score almost doubles and the CA centroid shifts from 14 $\mu$m to roughly 22 $\mu$m, beyond 4 pulses the curves flatten and subsequent gains are modest. Therefore, we consider that when the number of pulses is greater than or equal to 4, the organoid tends to produce more stable and more distinguishable outputs. For phase amplitude the three metrics increase gradually: the centroid shift moves from 25 $\mu$m at 4 $\mu$A to 32 $\mu$m at 12 $\mu$A, with little change thereafter, and the Silhouette curve shows a similar muted rise before plateauing. By contrast, phase duration yields a more sustained escalation across the entire range tested, raising the Silhouette score from close to 0 to 0.63 and displacing the centroid by almost 30 $\mu$m while steadily boosting global spike output.

Because pulse counts of four and above give more compact and separable center-of-activity clusters, we restrict the Braille encoding to four to ten pulses, a range that retains efficacy while avoiding overstimulation. Applying identical stimulation to different electrodes shifts the activity center by roughly 20 $\mu$m (Figure~8\ref{subfig:CA_dist}), showing that routing distinct tactile regions to distinct electrodes can convey spatial information. The shift direction, however, does not track electrode position in a simple one-to-one fashion, and response strength depends on the local network state around each electrode. That is, a region with stronger tactile intensity does not necessarily elicit a stronger response from the corresponding electrode. The organoid therefore imposes its own intrinsic mapping between input location and population response.

\subsection{Performance of Organoid in Braille Classification}

\begin{figure}[htbp]
    \centering

    \begin{minipage}[t]{0.5\textwidth}
        \centering
        \paneltag{subfig:classify_single}
        \includegraphics[width=\textwidth]{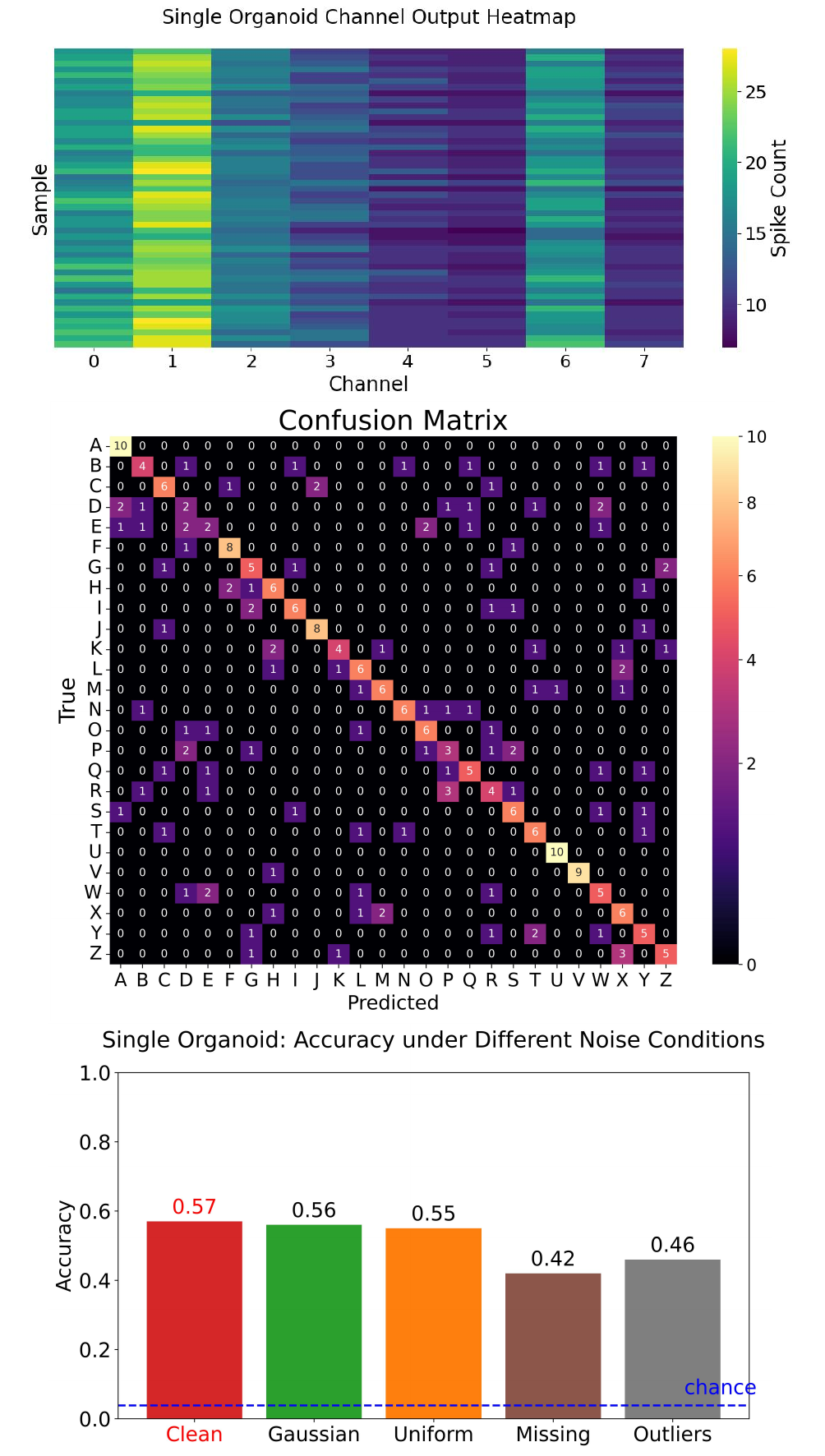}
    \end{minipage}\hfill
    \begin{minipage}[t]{0.5\textwidth}
        \centering
        \paneltag{subfig:classify_multi}
        \includegraphics[width=\textwidth]{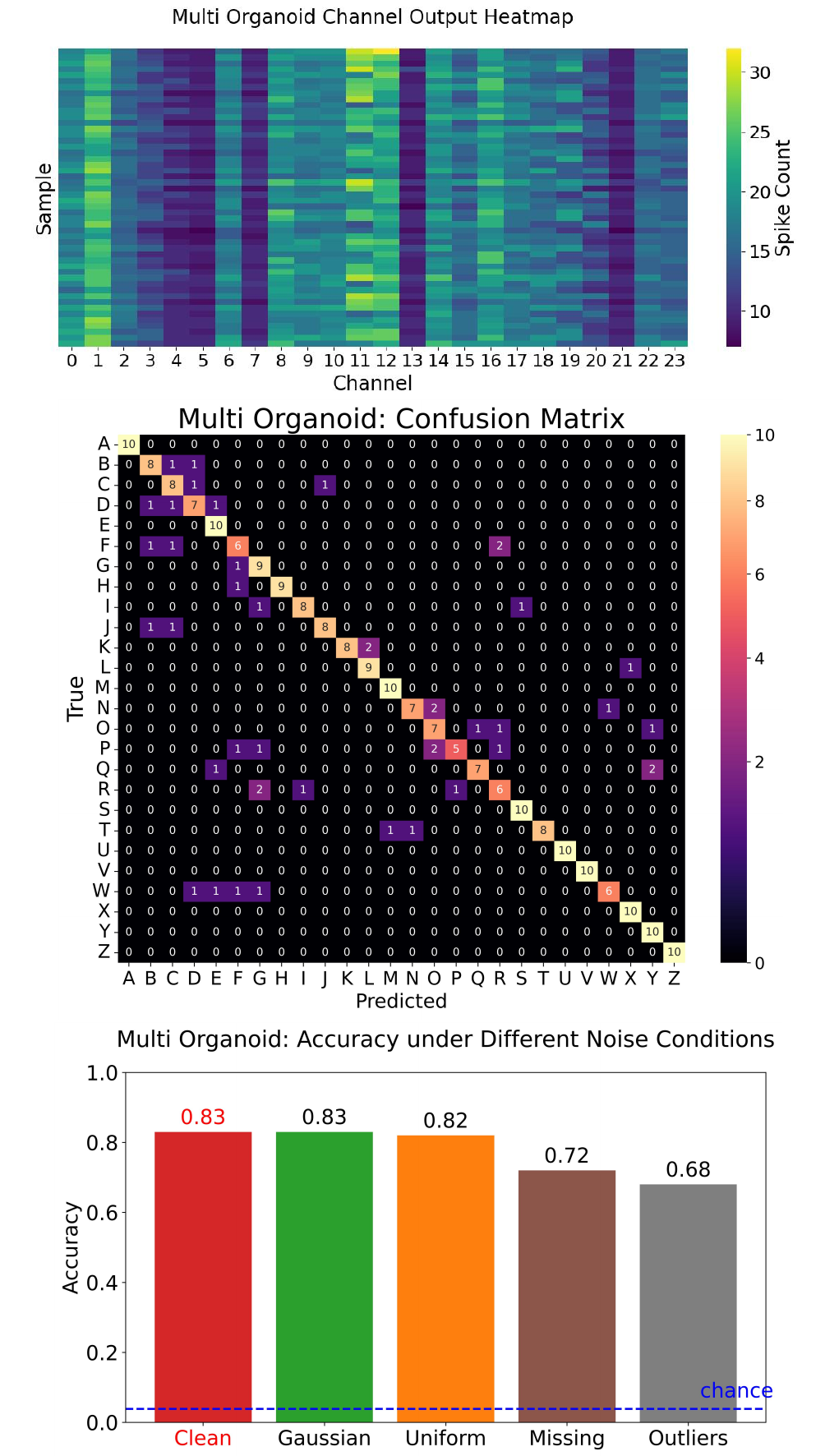}
    \end{minipage}

    \caption{Organoid output, Braille confusion matrix, and robustness to noise. (\subref{subfig:classify_single}) Results from using a single organoid. (\subref{subfig:classify_multi}) Results from using three organoids. From top to bottom: a heatmap of organoid output (each row corresponds to one tactile sample, each column to one electrode, with color indicating spike count); a confusion matrix showing classification of decoded organoid responses into 26 Braille letters using an SVM; and classification accuracy under different types of noise, where the red bar indicates performance without added noise. Chance level for 26 letters classification is 3.85\%.}
    \label{fig:braille_classification}
\end{figure}

As described previously in Section~\ref{sec:methodology_encoding}, tactile stimuli from the Evetac sensor are encoded into electrical stimulation patterns based on four parameters: phase amplitude, pulse count, phase duration, and trigger delay. Each tactile input is spatially partitioned into eight regions corresponding to the eight electrodes of the MEA. 
Organoid responses are decoded as trigger-aligned, threshold-based event counts from each electrode. For each sample, neural activity is recorded from 0 to 510 ms relative to trigger onset, the first 10 ms are excluded to suppress immediate stimulation artifacts~\citep{finalspark_closedloop_doc}, and counts from 10 to 510 ms are used as the 500 ms readout feature window. 
Here, we assess the effectiveness of this encoding and decoding strategy through the performance of the organoid system in classifying 26 Braille letters.

We use a support vector machine (SVM) classifier with cross-validation on 1300 response vectors for all classification analyses reported in this study.
When using only the 8-dimensional output from a single organoid for classification, the accuracy remains stable at approximately 57\% (Figure~9\ref{subfig:classify_single}). When conducting experiments with three organoids on the same MEA and using their individual 8-dimensional outputs separately for classification, the accuracies are 59\%, 62\%, and 67\%, respectively. However, when combining the outputs of the three organoids into a 24-dimensional vector, the classification accuracy using SVM reaches 83\% (Figure~9\ref{subfig:classify_multi}).

As shown in Figure~\ref{fig:braille_classification}, the heatmaps of organoid responses to each sample show that even though different pressing depths are used during data collection, the response patterns of the organoid remain relatively consistent,  
which indicates that the organoid network exhibits intrinsic robustness to variations in stimulus strength. Comparing the input and output heatmaps further confirms that a stronger indentation in a given sensor region does not always elicit a stronger response from the electrode assigned to that region. The transformation from tactile input to spike output is therefore shaped by local network dynamics rather than by a direct intensity mapping.
When noise is introduced again, the single-organoid system’s accuracy decreases by 2.1--26.5\% relative to its clean (no-noise) baseline, with 'missing' and 'outliers' noise causing the largest drops; the three-organoid system’s accuracy decreases by 0.8--18.1\% relative to its clean baseline. 
Here, the reported ranges are taken across the four noise models. 
Within the organoid-only comparison, these results show that combining multiple organoids enhances resistance to data disturbances. A direct comparison with decoding from the delivered stimulation pattern is presented in Section~\ref{subsubsec:Comparison_with_Direct_Decoding_from_the_Stimulation_Pattern}.

\subsection{Comparison with Direct Decoding from the Stimulation Pattern}
\label{subsubsec:Comparison_with_Direct_Decoding_from_the_Stimulation_Pattern}

To clarify the role of the organoid in the decoding process, we compared the three-organoid system with a direct stimulation baseline, in which the same SVM classifier was applied directly to the delivered stimulus waveform rather than to organoid responses.
This comparison was introduced to test whether the organoid stage weakens the task-relevant information already present in the encoded input, or whether it preserves that information in a useful biological substrate.

\begin{figure}[htbp]
    \centering

    \begin{minipage}[t]{0.42\textwidth}
        \centering
        \paneltag{subfig:classify_single}
        \includegraphics[width=\textwidth]{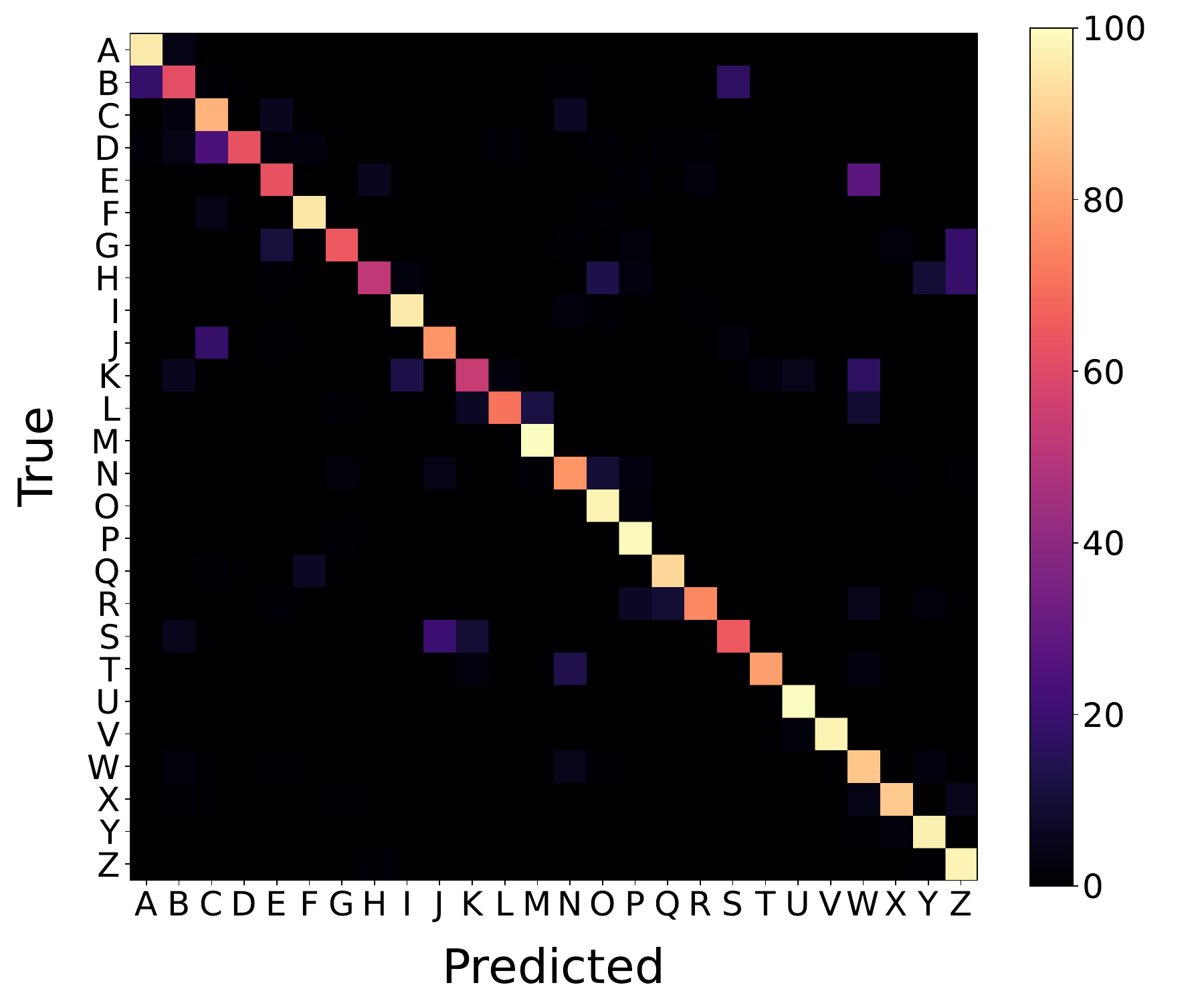}
    \end{minipage}\hfill
    \begin{minipage}[t]{0.53\textwidth}
        \centering
        \paneltag{subfig:classify_multi}
        \includegraphics[width=\textwidth]{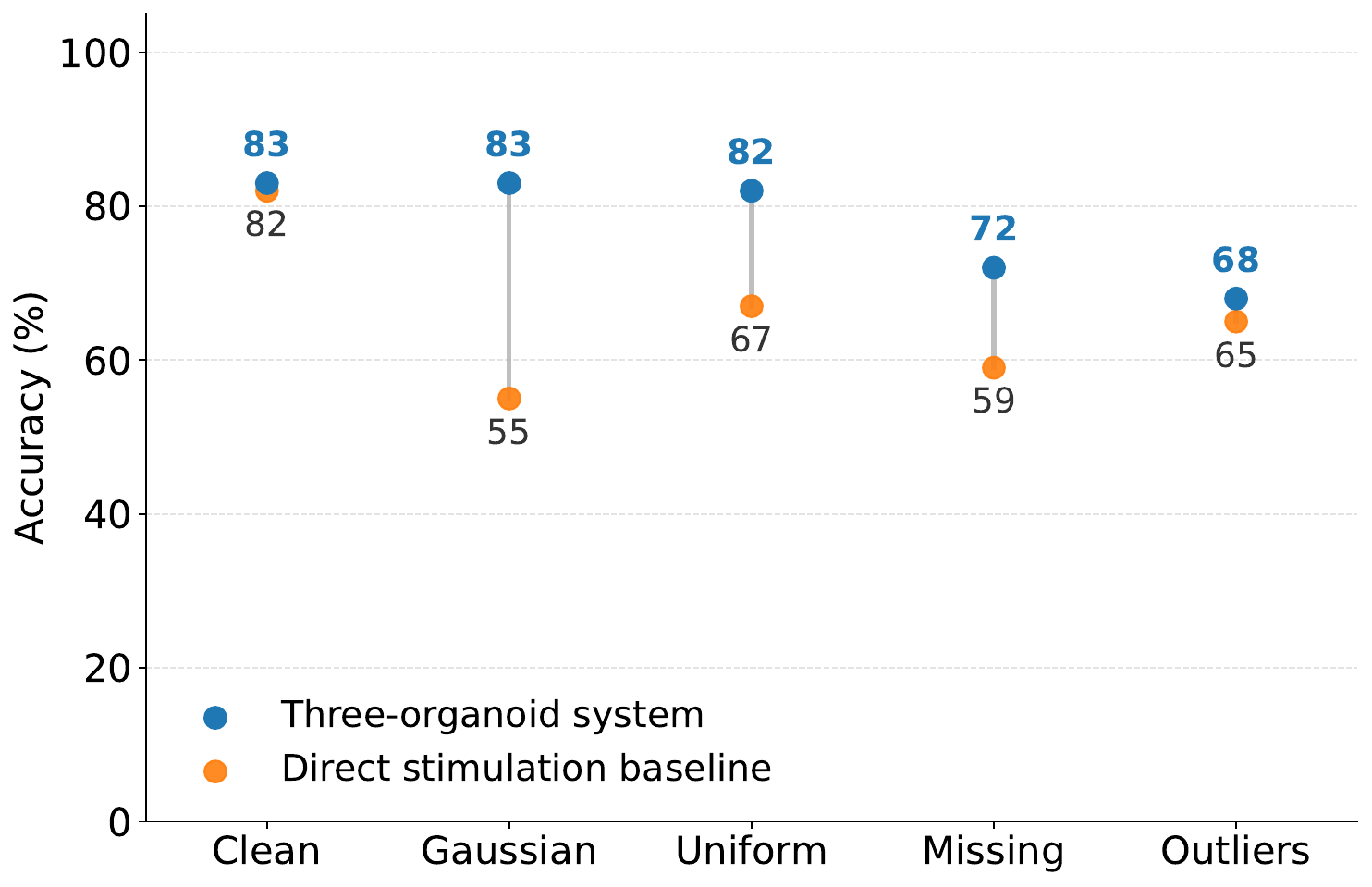}
    \end{minipage}

    \caption{Comparison with direct decoding from the stimulation pattern.
    (a) Confusion matrix obtained by applying the same SVM classifier directly to the delivered stimulation pattern.
    (b) Mean classification accuracy under clean and corrupted conditions for the three-organoid system and the direct stimulation baseline. The three-organoid system maintains comparable clean performance and shows stronger robustness under channel corruption.}
    \label{fig:baseline_comparison}
\end{figure}

As shown in Figure~\ref{fig:baseline_comparison}, the direct stimulation baseline achieved 82\% mean accuracy under clean conditions, indicating that the delivered stimulation pattern itself already contains substantial class-discriminative information. This is expected, since the encoding was designed to preserve task-relevant spatiotemporal cues from the tactile signal. The three-organoid system achieved 83\% under the same clean condition, showing that introducing the organoid stage does not reduce the decodable structure already present in the stimulation pattern. Instead, the multi-organoid configuration preserves that structure at a comparable level under clean decoding.

The difference becomes more evident under corruption. Using the same channel-corruption protocol described in Section~\ref{subsubsec:Organoid_Decoding_and_Classification_Under_Noisy_Conditions}, the direct stimulation baseline achieved 55\%, 67\%, 59\%, and 65\% under Gaussian noise, uniform noise, missing-data corruption, and outlier corruption, respectively. Under the same conditions, the three-organoid system achieved 83\%, 82\%, 72\%, and 68\%. Thus, while the clean performances are closely matched, the three-organoid system retains higher accuracy across all tested corruption settings. These results indicate that, in the present open-loop setting, the organoid stage is not a redundant intermediate step. Rather, its main contribution lies in preserving decodable task information while providing improved robustness when multiple organoids are combined.

\section{Discussion}

By systematically characterizing the stimulus-response profile of human forebrain organoids cultured on FinalSpark’s low-density eight-electrode MEA, we developed a transferable spatio-temporal-amplitude encoding strategy that converts spike-based outputs from a neuromorphic sensor into electrical-stimulation patterns that can be processed by the organoid network. 
Applied to a blind Braille-letter classification benchmark, the encoder demonstrated competitive performance, and multiplexing three organoids in parallel raised accuracy to 83\% while markedly improving robustness against Gaussian, uniform, data loss and outlier noise. 
Collectively, these results show that even sparsely instrumented organoids can support reliable open loop classification when supplied with appropriately structured stimulation input, laying essential groundwork for scalable bio-hybrid computing architectures.

\subsection{Toward a Transferable Encoding Strategy via Stimulus Parameter Optimization}

Increasing the number of pulses is the most direct way to evoke burst-type activity: trains of four or more pulses reliably boost spike counts, especially when combined with higher phase amplitude, which together determine overall stimulus strength. Phase duration is chiefly a temporal variable; it sets the length of each biphasic pulse, defines the integration window for incoming charge, and only secondarily modulates response magnitude. Trigger delay merely schedules when the train begins and does not alter the total charge delivered.
Our findings are consistent with the amplitude and pulse-width dependencies reported by Wagenaar et al.~\citep{wagenaar2004effective}; spike output rises steadily with larger amplitudes and longer pulses, displays a clear threshold, and can still be elicited by very brief stimuli. They also align with Li et al.~\citep{li2022stretchable}, who identified five pulses per train as minimally effective, while our data show that four pulses already provide robust network engagement. 
Although each electrode on the LD-MEA samples a micro-ensemble of neurons rather than single units, the array still offers enough resolution to expose these parametric trends, confirming that even low-density hardware can capture population-level stimulus–response rules. These four timing-and-strength parameters, combined with electrode location, form the basis of our spatio-temporal-amplitude encoding strategy.

Our proposed encoding strategy can be adapted to different tactile sensing tasks and sensor configurations. Spatial resolution can be flexibly adjusted along either spatial dimension depending on task requirements. For instance, tasks like static pressing or shape recognition, where no dominant sliding direction exists and tactile events are uniformly distributed, benefit from a symmetric spatial grid. Conversely, scenarios involving a clear sliding direction, such as Braille reading or texture scanning, are better served by increasing resolution along the movement axis to capture richer spatiotemporal dynamics. 
The contribution of the encoding strategy is not the fixed Braille-specific mapping itself, but the general interface principle it provides for tactile-to-organoid communication. Event-based tactile sensors generate high-dimensional asynchronous signals, whereas low-density MEA organoid platforms provide only a small number of stimulation channels. The proposed encoding addresses this mismatch by compressing region-wise tactile information into a compact set of MEA-compatible stimulation parameters while preserving three classes of task-relevant cues: spatial location, temporal structure, and contact intensity or variation. In the present implementation, spatial information is represented through region-to-electrode assignment, temporal information is represented through phase duration and trigger delay, and intensity-related information is represented through pulse count and phase amplitude. Because the ranges of these stimulation parameters are selected according to the organoid stimulus-response profile, the encoding is not only a sensor feature mapping, but also a biologically constrained interface design. This makes the framework useful for future robotic tactile tasks and organoid-based computing systems that require structured conversion from sensor streams to biological stimulation inputs.

However, the current LD-MEA configuration constrains our ability to detect subtle spatial activity patterns. Future studies should leverage high-density arrays such as BioCAM DupleX~\citep{BioCAMDupleX}, CMOS-MEA5000~\citep{CMOSMEA5000}, and MaxTwo~\citep{MaxTwo}, which, by virtue of their much larger electrode counts, enable pre-experimental screening of the most active or information-rich sites~\citep{cai2023brain}. 
Our current centre-of-activity analysis shows no systematic drift toward the stimulated electrode. 
Implementing such high-density MEAs would refine the spatial partitioning step, potentially uncovering finer spatial dynamics and more comprehensively characterizing the full parameter landscape.

\subsection{Braille Benchmark: Performance and Insights}

In the Braille classification experiment using organoid outputs, we compare the classification performance of 8-dimensional data from a single organoid and 24-dimensional data from three organoids using an SVM classifier. Under the single-organoid condition, the classification accuracy for 26 Braille classes is approximately 61\%, and the response patterns are relatively consistent across different pressing depths. When using three organoids, the accuracy increases to 83\%. From the perspective of the classifier, this is because increasing the data dimensionality tends to improve the linear separability of the data points when the additional features contain non-redundant class information~\citep{cover2006geometrical}.

From the biological perspective, we hypothesize that the improvement arises from partially complementary response transformations across the three organoids. Differences in network connectivity, developmental state, and ongoing network dynamics may cause the same encoded stimulus to evoke similar but non-identical response patterns. Neural population coding depends on the structure of response correlations rather than on feature number alone~\citep{franke2016structures}, while heterogeneity combined with an appropriate correlation structure can improve coding fidelity~\citep{da2014high}. Concatenating these responses may therefore provide the classifier with shared task-relevant information together with organoid-specific discriminative features.
As shown in the noise robustness experiments, the three-organoid system shows a smaller drop in accuracy after corruption is introduced. 
We hypothesize that, although the same proportion of output channels is perturbed in each condition, the concatenated responses of the three organoids retain complementary class-relevant information distributed across partially distinct response patterns.
Although this approach may seem redundant, it is commonly observed in biological systems, where degeneracy enables distinct neural circuits to substitute for one another~\citep{friston2003degeneracy}, allowing predictive feedback loops in the visual cortex to fill in occluded information~\citep{rao1999predictive} and contextual cues in speech to reconstruct missing phonemes~\citep{warren1970perceptual}, thereby safeguarding perception against damage or noise.

To further clarify the role of the organoid in this task, we compared the three-organoid decoder with direct decoding from the delivered stimulation pattern. The direct baseline is already highly informative, reaching 82\% under clean conditions when the full 32-dimensional stimulation representation is used. Even under this comparison, the 24-dimensional three-organoid system reaches 83\%, showing that the organoid stage preserves the task-relevant structure contained in the encoded stimulation. The contrast becomes clearer under corruption: the three-organoid system consistently retains higher accuracy than the direct baseline across Gaussian, uniform, missing-data, and outlier perturbations. This pattern suggests that the value of the organoid stage in the present open-loop setting lies in maintaining decodable information within a biological substrate while improving tolerance to channel corruption. 
This comparison provides further support for the hypothesis that partially distinct biological responses can stabilize decoding under perturbation, while direct measurements of inter-organoid response correlations are required to test this mechanism.
Additional organoids may provide further gains when they contribute informative responses that are only partially correlated with those already included. The benefit is expected to diminish as the responses become increasingly redundant, while weak or noisy organoids may add uninformative dimensions and increase the amount of training data required. The scaling beyond three organoids and its saturation point remain to be determined.

When working with organoid outputs, which are relatively low-dimensional, traditional machine learning methods such as SVM and KNN are preferable due to their small number of parameters and simple model structure. Compared with neural networks, which are more complex, have larger parameter sizes, and require longer training times, these methods offer clear advantages in energy efficiency~\citep{strubell2020energy}. 
However, for tasks that involve more complex features, higher-dimensional data, or stricter accuracy requirements, and where energy consumption is less constrained, deep learning methods could be applied~\citep{ahmadi2021robust}.
The present study relies on simple spike counts. Future work should evaluate temporal-pattern classifiers~\citep{kuchler2025engineered}, population-vector readouts, graph-based methods, and other decoders, both in open and closed loop, to determine which metrics best capture the computation performed by the organoid.

\subsection{Future Work}

The present open-loop experiment should be interpreted as an encoding and decoding study rather than as a direct test of organoid learning or long-term plasticity. Establishing whether feedback can induce adaptive changes in organoid network dynamics will require future closed-loop experiments. 
Neuromorphic tactile sensors and cerebral organoids both communicate through spikes. 
Although these spikes are converted into electrical pulses for stimulation, the encoding is designed to retain key spatiotemporal cues while reparameterizing sensor spikes into a stimulation space supported by the hardware interface.
Future work should investigate how the organoid networks interpret and transform these encoded spike patterns to engage their internal neural circuitry, potentially uncovering deeper insights into the organoids' intrinsic neural processing capabilities~\citep{osaki2024complex}.
Moreover, spike-derived stimulation could be further developed to facilitate clear channel-to-channel mapping, targeted neural activation, and biologically realistic starting points for adaptive, closed-loop bio-hybrid systems that exhibit learning and plasticity~\citep{ades2024biohybrid, habibollahi2023critical}.

In addition to spike-driven encoding, the present encoder uses a fixed 2 × 4 grid tuned to the Braille trajectory.  
Future work should explore varying the grid geometry, feature selection, and parameter quantisation to better understand how each factor influences classification accuracy, robustness, and energy efficiency.
A wider task set including static indentation, texture tracking and edge orientation detection could test whether specific input statistics preferentially engage organoid circuitry or even induce experience-dependent plasticity~\citep{robbins2024goal}.

Future work should systematically evaluate one-, two-, three-, and four-organoid configurations using the same channel-level corruption protocol and identical cross-validation splits. Pairwise response and classification-error correlations should be quantified to determine whether each additional organoid contributes non-redundant information. Organoid-ablation analyses should further measure the marginal contribution of each organoid and identify whether accuracy and robustness eventually reach a saturation point.

Future studies should also implement closed-loop platforms, allowing organoid-generated spikes to influence subsequent inputs actively. Comparing open-loop and closed-loop conditions will reveal if feedback mechanisms enhance learning efficacy and induce significant changes in network dynamics.
Such experiments can address the critical question of whether feedback is essential for organoid-based computation and adaptation~\citep{habibollahi2023critical}.
Finally, a forward-looking goal is to route organoid output into peripheral or cortical tissue, creating a bidirectional bio-hybrid loop in which synthetic sensors drive the organoid and the organoid modulates living nerves. Such preparations could serve as testbeds for neuro-prosthetic~\citep{oddo2016intraneural} concepts before in-vivo implementation.

\section*{Conflict of Interest Statement}

The authors declare that the research was conducted in the absence of any commercial or financial relationships that could be construed as a potential conflict of interest.

\section*{Author Contributions}

TL: Conceptualization, Methodology, Investigation, Formal analysis, Validation, Visualization, Writing – original draft;
HP: Conceptualization, Methodology, Writing – review \& editing;
BWC: Conceptualization, Supervision, Formal analysis, Writing – review \& editing.

\section*{Funding}

This paper is funded by the China Scholarship Council. This work was supported by the Engineering and Physical Sciences Research Council (https://ukri.org) grant number UKRI166.

\section*{Acknowledgments}
The authors thank FinalSpark for providing access to their Neuroplatform and neural organoid resources. We are grateful to the FinalSpark team for their technical support and valuable assistance during the experimental sessions.

\section*{Supplemental Data}
Not applicable.

\section*{Data Availability Statement}
The raw data supporting the conclusions of this article will be made available by the authors, without undue reservation.

\bibliographystyle{plainnat}
\bibliography{ref}

\end{document}